\documentclass[conference]{IEEEtran}
\IEEEoverridecommandlockouts
\usepackage{cite}
\usepackage{amsmath,amssymb,amsfonts}
\usepackage{algorithmic}
\usepackage{graphicx}
\usepackage{textcomp}
\usepackage{xcolor}

\ifCLASSOPTIONcompsoc 
\usepackage[caption=false,font=normalsize,labelfon t=sf,textfont=sf]{subfig} 
\else \usepackage[caption=false,font=footnotesize]{subfig} 
\fi

\def\BibTeX{{\rm B\kern-.05em{\sc i\kern-.025em b}\kern-.08em
    T\kern-.1667em\lower.7ex\hbox{E}\kern-.125emX}}

\usepackage{adjustbox}
\usepackage{amsfonts} 
\usepackage{amsmath}
\usepackage{bm}
\usepackage{enumitem}
\usepackage{dirtytalk}
\usepackage{multirow}
\usepackage{MnSymbol}
\usepackage{url} 
\usepackage{xspace}

\usepackage{hyperref}

\newcommand{\softmax}{\ensuremath{\mathrm{softmax}}}

\newcommand{\eg}{\emph{e.g.,}\xspace}

\newcommand{\etal}{\emph{et al.,\@}\xspace}

\begin{document}

\title{Predicting and Explaining Hearing Aid Usage Using Encoder-Decoder with Attention Mechanism and SHAP}

\author{\IEEEauthorblockN{
Qiqi Su}
\IEEEauthorblockA{\textit{Department of Computer Science} \\
\textit{City, University of London}\\
London, UK \\
qiqi.su@city.ac.uk}
\and
\IEEEauthorblockN{
Eleftheria Iliadou}
\IEEEauthorblockA{\textit{1st Department of Otorhinolaryngology} \\
\textit{Head and Neck Surgery, National}\\
\textit{and Kapodistrian University of Athens Medical School} \\
Athens, Greece \\
iliadoue@med.uoa.gr}
}

\maketitle

\begin{abstract}
It is essential to understand the personal, behavioral, environmental, and other factors that correlate with optimal hearing aid fitting and hearing aid users' experiences in order to improve hearing loss patient satisfaction and quality of life, as well as reduce societal and financial burdens. This work proposes a novel framework that uses Encoder-decoder with attention mechanism (attn-ED) for predicting future hearing aid usage and SHAP to explain the factors contributing to this prediction. It has been demonstrated in experiments that attn-ED performs well at predicting future hearing aid usage, and that SHAP can be utilized to calculate the contribution of different factors affecting hearing aid usage. This framework aims to establish confidence that AI models can be utilized in the medical domain with the use of XAI methods. Moreover, the proposed framework can also assist clinicians in determining the nature of interventions. 
\end{abstract}


\section{Introduction}

Unaddressed hearing loss (HL) is a major public health concern that places an enormous burden on societies and individuals (World Health Organization, 2017). It is estimated that only one in five individuals in need of a hearing aid (HAid) will acquire and continually use one. 

Currently, HAid models and configurations are selected based on functional hearing testing and patients' preferences for sound and lifestyle. Long-term use and a high level of benefit are indicative of a successful HAid fitting. They may be evaluated through the use of structured validated questionnaires (\eg Glasgow Hearing Aid Benefit Profile) or data logging (\eg hours spent using hearing aids). In adults, HAid usage is estimated to be 8 to 9 hours per day (Iwahashi \etal 2013; Kochkin, 2010; Laureyns \etal 2020). Daily and efficient use of HAids is associated with higher levels of satisfaction and benefit (Jilla \etal 2020; Singh \etal 2015). A better understanding of the factors that influence the use of HAids could lead to better fitting decisions. 

To date, the majority of studies assessing hours of usage and HAid benefit have been conducted through surveys or interviews with HAids wearers (Iwahashi \etal 2013; Kochkin, 2010; Laureyns \etal 2020). Although very informative and useful for gaining an initial insight, this research design has particular limitations (Andrade, 2020). First of all, data are being collected from a conveniently available pool of responders – participants most of the time (\eg patients of a particular Audiology Clinic or residents of a specific area). This convenience sampling is time-efficient and can be helpful when targeting a specific population, however it may also adversely affect the generalizability and reproducibility of the findings. In addition, self-reported information may be significantly affected by participants’ memory (recall bias) or fatigue (non-response or acquiescence bias). Especially in the case of HAids, previous studies have proven that patients tend to overreport their daily hours of HAid use (Laplante-Lévesque \etal 2006). Finally, researchers’ methods of collecting, recalling, recording, or handling of information may also affect findings (information bias). Therefore, a high priority is placed on developing new methodologies that overcome the aforementioned limitations and are based on real-world data related directly to the use of the HAid.

\section{Related Work}

In recent years, Artificial Intelligence (AI) models have gained increasing traction in the HL prognosis domain. Among these models, recurrent-based Deep Learning (DL) models such as Long Short-Term Memory (LSTM) (Hochreiter \& Schmidhuber, 1997) are particularly popular due to their ability to handle the sequential properties found in many HL-related datasets, as well as capable of dealing both long-term and short-term time series (Preeti \etal 2019). LSTM has been applied to a number of HL-related fields, including speech assessment (Chiang \etal 2021), speech enhancements (Garg, 2022; Zhang \etal 2019), as well as daily routine recognition using acceleration and audio data from the HAids (Kuebert \etal 2021).

In order to further enhance the performance of LSTMs, many researchers have proposed a variety of methods. In particular, the combination of LSTMs and Attention Mechanisms has become increasingly popular, as it is often able to provide better results, especially for sequence problems (Liu \& Guo, 2019). With an Encoder-Decoder (ED) architecture, Bahdanau \etal (2014) first introduced the attention mechanism in a Neural Machine Translation task. The use of attention mechanisms in conjunction with EDs or LSTMs can prevent the model from putting too much weight on certain input features, as well as enable the model to obtain correlations between input and target features (Xu \etal 2015). The application of attention mechanisms with ED in the hearing domain is currently limited to speech-related studies, such as speech enhancements (Lan \etal 2020) and electroencephalography responses (Lu \etal 2021). 

Despite the advancements of these state-of-the-art techniques however, the majority of the research in the HL prognosis domain still relies on traditional Machine Learning methods for prediction (Abdollahi \etal 2018; Bing \etal 2018; Lenatti \etal 2022; Tomiazzi \etal 2019; Zhao \etal 2019). This is partly because the integration of AI models in the medical domain still faces criticism for not adhering to the high standards of accountability, reliability, and transparency (Anderson, 2018). AI models are considered to be a black box where humans are not able to easily comprehend how models work or how they arrive at conclusions. In order to overcome this limitation, Explainable AI (XAI) techniques aim to explain the learned decisions to the end-user so they can trust the model, thus enabling them to use AI models in sensitive areas, such as the medical domain. While several XAI methods have been proposed over the years, very few studies have explored their potential applications in the medical fields (Tjoa \& Guan, 2021), and even fewer in the hearing domain. One such approach is the application of SHapley Additive exPlanations (SHAP) (Lundberg \& Lee, 2017) to explain the classification results of a Random Forest model in predicting the HL status of a patient (Lenatti \etal 2022).  

Therefore, in this study, we propose a novel AI-XAI framework that combines attn-ED and SHAP to analyze the impact of parameters such as age and average pure tone thresholds on the prediction of hours of HAid use per day in adults with mild to moderate HL. This is the first framework, to the best of our knowledge, that utilizes the predictive power of AI models as well as the explanatory capabilities of XAI methods specifically for HAid usage and benefit. 

The rest of the paper is as follows. Section 3 describes the dataset, the pre-processing steps, and the proposed framework. Results of the proposed framework in providing a personalized and a global solution are presented in Section 4. Section 5 discusses the paper and finally, Section 6 concludes the paper.  
 
\section{Methods}

\subsection{Dataset}

A synthesized sample of the EVOTION dataset with 399,500 observations of 53 participants is used in this study (Christensen \etal 2019a). The EVOTION project recruited almost 1,000 participants with varying degrees of HL, and each participant was supplied with a pair of EVOTION HAids that were connected to a smartphone by low-energy Bluetooth (Christensen \etal 2019b). The smartphone logged real-time data every minute that included both smartphone and HAids data (Christensen \etal 2019b). The dataset consists of various variables, such as: ID, HAid setting, sound environment, location, timestamps, acoustic parameters, and the degrees of HL on the best hearing ear of the participant. 

It is considered necessary to include some personal characteristics of each participant that are missing from the sampled EVOTION data in order to accomplish the objective of study, which is to identify characteristics that make participants more inclined to use their HAids sufficiently long during the day. Therefore, two additional variables, which are regularly collected by clinicians during audiologic clinical evaluations – Sex and Age – are randomly populated and added to the dataset. Furthermore, acoustic parameters are excluded from the sampled EVOTION data since, according to our understanding, these are HAid model specific and would affect the interpretation and generalizability of our results. \tablename~\ref{tab: dataset} summarizes the variables with their description and type.

\begin{table}
    \centering
    \caption{LIST OF VARIABLES WITH THEIR DESCRIPTION AND TYPE IN THE EVOTION SAMPLED DATASET}
    \begin{adjustbox}{width=\linewidth}
    \begin{tabular}{ccc}
        \hline
         Variable Names & Description & Type\\
         \hline
         \hline
         \\[-1em]
         ID	& Identifier & Integer \\
         \\[-1em]
         \hline
         \\[-1em]
         Age & Years of age & Integer \\
         \\[-1em]
         \hline
         \\[-1em]
         \multirow{2}{*}{Sex}
         & Biological gender: & Categorical \\
         & female and male & \\
         \\[-1em]
         \hline
         \\[-1em]
         \multirow{3}{*}{hProg}
          & HAid program: & Categorical \\
          & low, medium,  & \\
          & high, and high+ & \\
          \\[-1em]
         \hline
         \\[-1em]
         hVol & HAid volume & Integer \\
         \\[-1em]
         \hline
         \\[-1em]
         LatRel & Relative latitudinal & Continuous \\
         \\[-1em]
         \hline
         \\[-1em]
         LonRel & Relative longitudinal & Continuous \\
         \\[-1em]
         \hline
         \\[-1em]
         \multirow{4}{*}{PTA4}
          & Pure tone average & Continuous \\
          & across 4 testing & \\
          & frequencies (0.5, 1,  & \\
          & 2, and 4kHz) & \\
          \\[-1em]
         \hline
         \\[-1em]
         \multirow{4}{*}{SoundClass}
          & Sound environment: & Categorical \\
          & quiet, speech, & \\
          & speech in noise, & \\
          & and noise & \\
          \\[-1em]
         \hline
         \\[-1em]
         Timestamp & Time of the logged record & ISO 8601 \\
         \\[-1em]
         \hline
    \end{tabular}
    \end{adjustbox}
    \label{tab: dataset}
\end{table}

\subsection{Analysis Methods}

The proposed framework aims to use a predictive model – ED with attention mechanism (attn-ED) – to predict participants’ future daily HAids usage, then identify the characteristics that influence the model prediction through SHAP. The problem to be solved by the attn-ED is therefore a forecasting one. More specifically, it is a time series forecasting problem as the model is fundamentally constructed based on participants’ historical HAids usage and habit. Using the proposed framework, clinicians also have the freedom to choose between personalized predictions and explanations, and a global one. 

\subsubsection{Preprocessing the Data}

To calculate the total HAids usage in seconds per day per patient, the usage interval is calculated first. The usage interval in seconds is obtained from the Timestamp variable. The term, Distance, is defined first which is the difference in seconds between two consecutive timestamps, $t_i$ and $t_{i+1}$, of when the measurements are taken, such that:

\begin{equation}
    D = t_{i+1} - t_i
\end{equation}

The maximum distance, $D_max$, between two consecutive timestamps is set to 600 seconds in this analysis. Therefore, two consecutive measurements $m_i$ and $m_{i+1}$ taken at $t_i$ and $t_{i+1}$ respectively, belong to the same interval, $u_t$, if and only if the distance between the two timestamps is less or equal to 600 seconds, such that $|D| \leq {D_max}$. $m_{i+1}$ therefore, belongs to the subsequent interval, $u_{t+1}$, if the distance is greater than 600 seconds. The Usage Interval Duration, $d_t$, of each $u_t$ is then calculated as:

\begin{equation}
    d_t = t_e - t_s
\end{equation}

where $t_e$ is the last (maximum) timestamp of $u_t$ and $t_s$ is the first (minimum) timestamp of $u_t$. Finally, the HAid Usage, $h_t$, per day is calculated by taking $d_1 + d_2 + \ldots + d_n$ for all intervals in a day. 

For other variables to be transformed with a daily frequency, the average of each variable is taken for each day for each patient. Since hProg and Soundclass are categorical variables, these are transformed with ordinal encoding first before the aggregation. Whereas Sex is transformed with one-hot encoding instead. Continuous variables are scaled with standardization and normalization to ensure the range of data lies between a smaller range so that the model can learn better. 

Handling missing data is an important preprocessing step to be taken since the percentage of missing data in medical data can be as high as 98\% (Chan \etal 2010). As a result, simply deleting the rows of missing data is not feasible here. Aside from missing variable values in the dataset, there are also missing timestamp values resulting in gaps in the time series. As the data used in this study is longitudinal, meaning that the same variable is repeatedly measured at different times, Trajectory Mean method is suitable here (Genolini \etal 2013) where the mean of the observed values per participant is imputed. 

Outliers are defined as \say{samples that are exceptionally far from the mainstream data} (Kuhn \etal 2013). Since the standard deviation method is more suitable to detect outliers in data with Gaussian or Gaussian-like distributions (Ilyas \& Chu, 2019), it is used in this study after the dataset is standardized and normalized. The z-score for every $\xi_i$, which is the number of standard deviations away from the mean is calculated first. Data points can be declared as outliers if their z-score standard deviation is greater than a predefined threshold, which is set to three in this study. 

A Variance Inflation Method (VIF) is then used to determine if there is a multicollinearity among the variables. Multicollinearity refers to when an independent variable is highly correlated with one or more of the other independent variables in a regression model (Allen, 1997). A VIF value of 10 in general indicates that there is a weak multicollinearity between each independent variable, and those variables with a VIF value of less 10 can be included in the model. 

Finally, the data is split into training, validation, and testing sets. In particular, the splitting takes place for each participant, where 80\% of the data per participant is used as the training set, 20\% of the training set is used as the validation set, and 20\% of the data is reserved as the unseen testing set. All training, validation, and testing sets are then merged together so that the model can be trained with global data yet still be able to make personalized predictions, as well as global predictions.

\subsubsection{Proposed Model Architecture} 

\begin{figure}[t]
    \centering
    \includegraphics[width=1\linewidth]{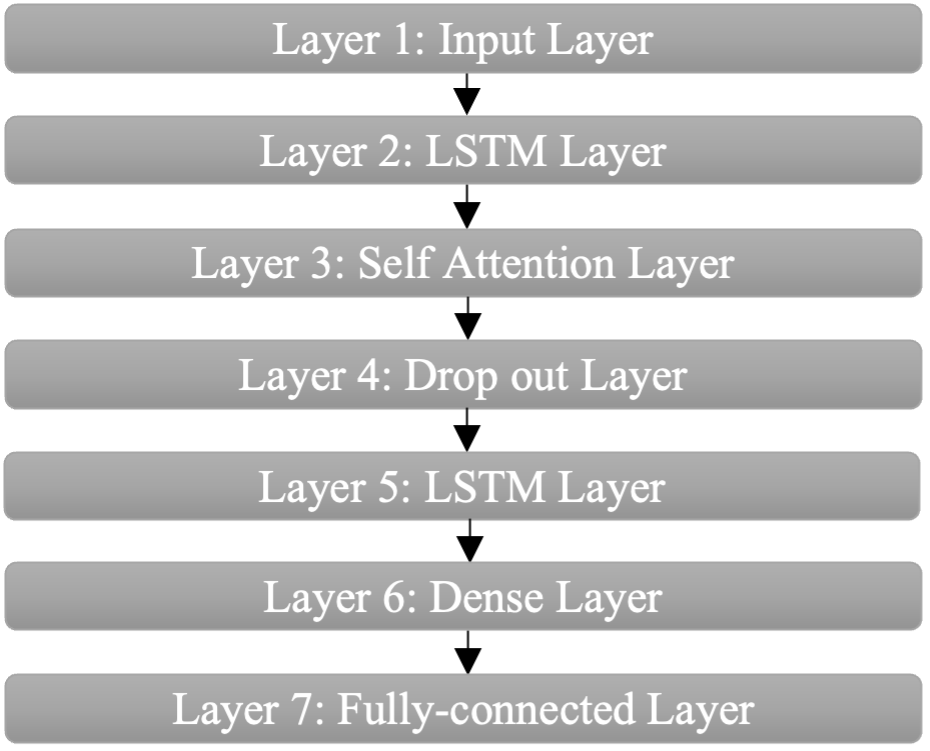}
    \caption{The architecture of the proposed model – attn-ED}
    \label{fig: architecture}
\end{figure}

Figure~\ref{fig: architecture} illustrates the architecture of the proposed model. In this study, the encoder uses an LSTM to encode the input sequences into a vector with fixed dimensionality, while the decoder uses another LSTM to decode the target sequences from the fixed vectors. An LSTM is a refined variation of the Recurrent Neural Network (RNN) that overcomes the potential problem of gradient vanishing that is a common occurrence for RNNs. An LSTM contains some recurrently connected special units called memory cells and their corresponding gate units in the hidden layer of an LSTM network. These three gates are input gate, forget gate, and output gate (Hochreiter \& Schmidhuber, 1997). The final hidden state, $h_t$, is calculated with a series of gating mechanism as follows: 

\begin{equation}
\left\{
\begin{aligned}
 f_t & = \sigma(w_a a_t + w_h h_{t-1} + b) \\
 i_t & = \sigma(w_a a_t + w_h h_{t-1} + b) \\
 s_t & = \tanh(w_a a_t + w_h h_{t-1} + b) \\
 c_t & = f_t \bigodot c_{t-1} + i_t \bigodot s_t \\
 o_t & = \sigma(w_x x_t + w_h h_{t-1} + b) \\
 h_t & = \tanh(c_t) \bigodot o_t
\end{aligned} \right.
\end{equation}

where $i_t$, $f_t$, and $o_t$ are the input gate, forget gate, and output gate, respectively. $x_t$ is the input vector at current timestep $t$. $w_x$ and $w_h$ are the weighting factor and $b$ is the bias vector. Furthermore, $\sigma$ represents the sigmoid function, tanh is the hyperbolic tangent function, and $\bigodot$ is the element-wise product. Lastly, $c_t$ denotes the cell state and $s_t$ is a newly created vector during the computation which decides if the new information should be stored in the cell state or not. 

The specific attention mechanism employed here is the Self-attention similar to the one proposed by Vaswani \etal (2017). Self-attention, also known as intra attention, aims to relate different positions of an input sequence in order to compute the representations of the same sequence. The inputs of the attention mechanism are therefore the sequence of hidden state vectors for all timesteps produced by the encoder LSTM, $H=(h_1, h_2, \ldots, h_n)$. A compatibility score of each hidden state vector in $H$ against the hidden state vector for which the self-attention is calculated first, as follows:

\begin{equation}
    \text{Compatibility Score} = \frac{HH^T}{\sqrt{d_H}}
\end{equation}

where $d_H$ is the dimension of the sequence of hidden state vectors. Finally, the output of the attention mechanism is calculated as a weighted sum of the hidden state vectors and the compatibility score. The matrix of the output is calculated as follows:  

\begin{equation}
    \textit{Attention}(H) = \softmax (\frac{HH^T}{\sqrt{d_H}}) H
\end{equation}

\subsubsection{Training the Model}

To optimize the prediction performance of the model, the hyper-parameters are hyper-tuned first with the validation set. \tablename~\ref{tab: hyp} summarizes the hyper-parameters to be tuned and their corresponding settings.  

The attn-ED is then trained on the training set with the set of optimal hyper-parameters obtained from hyper-tuning. The model is trained with 500 epochs with early call-back to avoid overfitting. 

The results of the trained attn-ED are reported by predicting the unseen testing set and evaluated using three standard error estimators:

\begin{equation}
    sMAPE = \frac{200}{N} \sum^N_{t=1} \frac{|y_i - \Tilde{y_i}|}{|y_i| + |\Tilde{y_i}|},
\end{equation}
\begin{equation}
    MAPE = \frac{1}{N} \sum^N_{t=1} \frac{|y_i - \Tilde{y}_i|}{y_i},
\end{equation}
\begin{equation}
    WAPE = \frac{\sum_{i=1}^N |y_i - \Tilde{y_i}|}{\sum_{i=1}^N |y_i|},
\end{equation}

where $y_i$ is the true value, $\Tilde{y}_i$ is the predicted value, and $N$ is the number of data points. 

\begin{table}[t]
    \centering
    \caption{HYPER-PARAMETERS AND THEIR CORRESPONDING SETTINGS}
    \begin{tabular}{cc}
        \hline
         Hyperparameteres & Settings \\
         \hline
         \hline
         \\[-1em]
         Number of LSTM hidden units & 32, 64, 128, 216, 512\\
         \\[-1em]
         Dropout rate in the LSTM layer & 0.01-0.0001\\
         \\[-1em]
         Recurrent dropout rate in the LSTM layer & 0.01-0.0001\\
         \\[-1em]
         Dropout layer  & 0.01-0.0001\\
         \\[-1em]
         Activation function in the fully-connected layer & ReLu, Sigmoid, Tanh\\
         \\[-1em]
         Learning rate & 0.01-0.0001\\
         \\[-1em]
         Batch size & 16, 32, 64\\
         \\[-1em]
         \hline
    \end{tabular}
    \label{tab: hyp}
\end{table}

\subsubsection{Explaining the Model}

SHAP (Lundberg \& Lee, 2017), more specifically, Kernel SHAP, is used to explain the prediction results of attn-ED in this study. SHAP assigns importance to each feature according to the Shapley values from Game Theory and aims to explain the predictions by computing the contribution of each feature to the model predictions. Mathematically, the SHAP explanations can be written as follows:

\begin{equation} \label{eq: shap_eq}
    g(z') = \phi_0 + \sum^M_{i=1} \phi_i z'_i
\end{equation}

where $g$ is the explanation model of the prediction model, $z' \in {0,1}^M$ where $z'$ is the binarized feature and $M$ is the number of binarized input features, $\phi_0$ is the model output without binarised inputs, and $\phi_i \in R$ are the Shapley values (Lundberg \& Lee, 2017).

SHAP is a local XAI method meaning that the method is designed to explain only the model prediction on a single data instance. However, it is also possible to obtain a global explanation with SHAP through aggregation, by calculating the mean absolute SHAP values for each feature across the dataset. Therefore, the relative impact of all features over the entire dataset and the global importance of each feature can be determined. 

The results of SHAP are presented in the form of a visualization using the SHAP Summary Plot as it combines the feature importance with feature effects. 

\section{Results}

The VIF value of each variable indicated that all variables can be included in the model in this study as they have a weak multicollinearity and are moderately correlated. There is a total of 11,643 datapoints after preprocessing the dataset, of which the training set contains 9,705 datapoints, the validation set contains 2,425 datapoints, and the testing set contains 1,938 datapoints. The optimal hyper-parameter settings are listed in \tablename~\ref{tab: opt_hyp}. 

\begin{table}[t]
    \centering
    \caption{OPTIMAL HYPER-PARAMETER SETTINGS AFTER HYPER-TUNING}
    \begin{adjustbox}{width=\linewidth}
    \begin{tabular}{cc}
    \hline
    Optimal Hyper-parameters & Settings \\
    \hline
    \hline
    \\[-1em]
    Number of LSTM hidden units & 128 \\
    \\[-1em]
    Dropout rate in the LSTM layer & 0.0002 \\
    \\[-1em]
    Recurrent dropout rate in the LSTM layer & 0.0041 \\
    \\[-1em]
    Dropout layer & 0.0008 \\
    \\[-1em]
    Activation function in the fully-connected layer & Tanh \\
    \\[-1em]
    Learning rate & 0.0013 \\
    \\[-1em]
    Batch size & 32 \\
    \\[-1em]
    \hline
    \end{tabular}
    \end{adjustbox}
    \label{tab: opt_hyp}
\end{table}

In order to benchmark the performance of attn-ED, the results of a Vanilla LSTM in predicting the dataset are also presented here. The architecture of a Vanilla LSTM just consists of an input layer, a single LSTM layer, and a fully-connected layer for making the prediction. All hyper-parameters are set to the default setting, with the number of hidden units set to 32 and no regularization. 

\subsection{Personalised Prediction and Explanation Results}

The results of training the models on all available data and predicting the future 14 days of HAids usage of only Participant 17 are presented in this section. \tablename~\ref{tab: personalised_preds} shows that attn-ED outperformed the Vanilla LSTM in all error estimators. It is worth to note that the reason why values of MAPE are relatively higher than other error estimators is because there are zero values in the test set, and MAPE tends to blow up when variable values are low. 

\begin{table}[!h]
    \centering
    \caption{ERROR ESTIMATORS FOR ATTN-ED AND VANILLA LSTM FOR PERSONALISED PREDICTION}
    \begin{adjustbox}{width=\linewidth}
    \begin{tabular}{ccc}
    \hline
    Error Estimators & attn-ED & Vanilla LSTM \\
    \hline
    \hline
    \\[-1em]
    sMAPE &	0.2460 & 1.2158 \\
    \\[-1em]
    MAPE & 0.7576 & 12.3022 \\
    \\[-1em]
    WAPE & 0.2513 & 1.0094 \\
    \\[-1em]
    \hline
    \end{tabular}
    \end{adjustbox}
    \label{tab: personalised_preds}
\end{table}
  
\begin{figure}[t]
    \centering
    \includegraphics[width=1\linewidth]{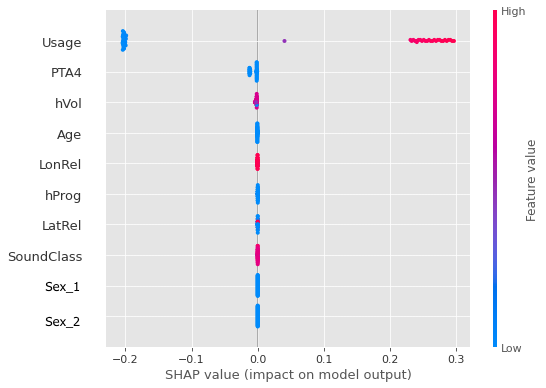}
    \caption{SHAP explanation result for the personalised prediction of Participant 17}
    \label{fig: personalised_shap}
\end{figure}

Figure~\ref{fig: personalised_shap} shows the SHAP explanation results for attn-ED making the personlized predictions. The x-axis of the plots represents the SHAP value, or the impact on the model prediction, of each feature, the y-axis lists all the features and ordered according to their importance, and the color depicts the value of the feature from low to high.   

The figure shows that Usage is the most important feature contributing to the prediction for Participant 17, with all other features make a negligible contribution to the prediction. For example, a lower value of PTA4 only negatively affects the model prediction marginally. Furthermore, a higher daily HAids usage corresponds to a higher future usage, while a lower HAid usage leads to a lower future usage. 

\subsection{Global Prediction and Explanation Results}

The results of training the model on all available data and predicting the future 14 days of HAids usage of all participants are presented in this section. \tablename~\ref{tab: global_preds} shows that attn-ED also outperformed Vanilla LSTM in all error estimators.  

\begin{table}[t]
    \centering
    \caption{ERROR ESTIMATORS FOR ATTN-ED AND VANILLA LSTM FOR GLOBAL PREDICTION}
    \begin{adjustbox}{width=\linewidth}
    \begin{tabular}{ccc}
    \hline
    Error Estimators & attn-ED & Vanilla LSTM \\
    \hline
    \hline
    \\[-1em]
    sMAPE & 0.1394 & 0.4588 \\
    \\[-1em]
    MAPE & 1.7247 & 7.4789 \\
    \\[-1em]
    WAPE & 0.1572 & 0.4019 \\
    \\[-1em]
    \hline
    \end{tabular}
    \end{adjustbox}
    \label{tab: global_preds}
\end{table}

\begin{figure}[t]
    \centering
    \includegraphics[width=1\linewidth]{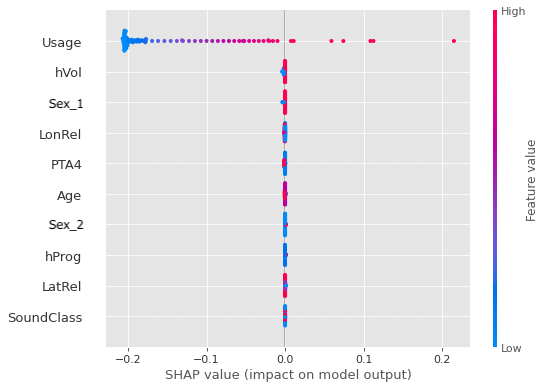}
    \caption{SHAP explanation result for the global prediction}
    \label{fig: global_shap}
\end{figure}

It is also observed that higher daily HAid usage leads to higher future usage among all participants, as shown in Figure~\ref{fig: global_shap}. While all other features contribute very little to the model prediction, hVol is the second most important feature contributing the model prediction for all participants, and SoundClass is the least important feature. 

It is also worth to note that Sex\_1 corresponds to Female and Sex\_2 corresponds to Male. Therefore, Figure~\ref{fig: global_shap} also indicates that female participants contribute more to the model prediction than male participants. 

\section{Discussion}

An increase in the time that HAid users actively use their HAid can have a positive impact on their HAid experience, communication, and overall quality of life. The goal of this study is to understand these parameters that may relate to this increase. The results of the proposed framework have shown that it is capable of producing good prediction results as well able to explain the predictions. Furthermore, the proposed framework can also be useful to clinicians if they wish to obtain a relatively personalized solution for the participants. 

An important contribution of this paper is the use of XAI methods to assist clinicians in gaining a deeper understanding of factors that influence a participant's use of HAids. Previous studies have correlated longer HAid use with improved HAid experience and overall benefit (Houmøller \etal 2022; Jilla \etal 2020; Kaplan-Neeman \etal 2012). Our analysis showed that the use of HAids for sufficiently long periods of time relates directly with higher HAid future use, as well. 

Our finding may have various explanations and clinical implications. First of all, it is likely that those participants who have already adapted using HAids to their daily lives, who have integrated their HAids more efficiently into their communication strategies and behaviors, and thus already use them for a longer period of time will be using them more frequently. As an important factor in motivating participants to continue using their HAids, this information may be valuable if included in Audiologist consultation - counselling. Furthermore, initial low HAid usage due to suboptimal fittings or ineffective counselling may require the intervention of an audiologist in order to prevent future low HAid usage. Finally, further investigation of the timeframe that is significant for defining each participant’s future HAid usage, and experience may allow for the development of an evidence-based and personalized HAid counseling program, thereby optimizing the current empirical system, and eventually saving patients' time and resources.

According to our results, active hearing aid volume state (compared to default volume) and users’ PTA4 are also weakly associated with future HAid usage. HAid volume changes may be interpreted as a suboptimal fit according to the user’s daily life conditions and preferences, or, on the contrary, as the user’s capacity to handle their HAid manually and adapt it to any situation. Additionally, PTA4 may indicate that people with better hearing may benefit more from their HAid and will thus be able to utilize it more effectively and longer. Even so, it is important to remember that PTA4 only accounts for a small fraction of one's hearing phenotype, and the relationship between audiometric data and HAid usage, and experience is much more complex. It is warranted to conduct more audiometric, neurophysiological, and electrophysiological hearing studies in the future. 

One limitation of the proposed framework is that it lacks evaluations for the XAI method. Although the proposed framework is validated by clinicians, this is only a subjective assessment of the XAI method. Although there are currently no widely accepted objective metrics for evaluating XAI methods, existing metrics such as Rosenfield's set (Rosenfeld, 2021) should be tested in the future in order to obtain both objective and subjective validation. Another important limitation is that the used data are originating from a single source of synthetic data. As a consequence, important data that are relevant to HAid usage and benefit, such as patients’ experience (first-time users vs experienced users), audiometric data other than PTA4, user’s education level, or social activity and cognition level are missing. In the future, we are aiming at implementing the current approach to a large real-world dataset including a wide range of heterogenous data (Iliadou \& Su, 2022).

\section{Conclusion}

HAid experience is a complex problem to be solved, and this model is an initial exploration in this domain. Therefore, the proposed framework will be used to answer more questions relating to HAids benefit and usage in the future, such as identification of those factors that make participants more prone to stop using their HAids or augment the benefit of participants from using their HAids. The proposed framework is not limited to the hearing domain and will be interesting to evaluate its effectiveness for other comorbidities.

We believe that the proposed framework will be beneficial for multiple stakeholders in the hearing domain, such as clinicians, HAids users, researchers, and healthcare policy makers. Knowledge gained through this study can also be applied to the analysis of non-synthetic data. In the long run, the goal is the adaptation of counselling and model selection and fitting, as well as consideration in the design of future public health policies. It is also expected that the proposed framework will increase confidence in the use of AI and XAI methods in the medical domain. 

\section*{Acknowledgment}

This work was supported by the European Commission's Horizon 2020 research and innovation program under the SMART BEAR project (Grant agreement number: 8557172/H2020-SC1-FA-DTS-2018-2.) 


\section*{References}

\noindent [1]	Abdollahi, H., Mostafaei, S., Cheraghi, S., Shiri, I., Rabi Mahdavi, S., \& Kazemnejad, A. (2018). Cochlea CT radiomics predicts chemoradiotherapy induced sensorineural hearing loss in head and neck cancer patients: A machine learning and multi-variable modelling study. \textit{Physica Medica, 45}, 192–197. \url{https://doi.org/10.1016/j.ejmp.2017.10.008}

\noindent [2]	Allen, M. P. (1997). The problem of multicollinearity. In \textit{Understanding Regression Analysis} (pp. 176–180). Springer US. \url{https://doi.org/10.1007/978-0-585-25657-3_37}

\noindent [3]	Anderson, C. (2018). Ready for Prime Time?: AI Influencing Precision Medicine but May Not Match the Hype. \textit{Clinical OMICs, 5(3)}, 44–46. \url{https://doi.org/10.1089/clinomi.05.03.26}

\noindent [4]	Andrade, C. (2020). The Limitations of Online Surveys. \textit{Indian Journal of Psychological Medicine, 42(6)}, 575–576. \url{https://doi.org/10.1177/0253717620957496}

\noindent [5]	Bahdanau, D., Cho, K., \& Bengio, Y. (2014). Neural Machine Translation by Jointly Learning to Align and Translate. \url{ArXiv:1409.0473}.

\noindent [6]	Bing, D., Ying, J., Miao, J., Lan, L., Wang, D., Zhao, L., Yin, Z., Yu, L., Guan, J., \& Wang, Q. (2018). Predicting the hearing outcome in sudden sensorineural hearing loss via machine learning models. \textit{Clinical Otolaryngology, 43(3)}, 868–874. \url{https://doi.org/10.1111/coa.13068}

\noindent [7]	Chan, K. S., Fowles, J. B., \& Weiner, J. P. (2010). Review: Electronic Health Records and the Reliability and Validity of Quality Measures: A Review of the Literature. \textit{Medical Care Research and Review, 67(5)}, 503–527. \url{https://doi.org/10.1177/1077558709359007}

\noindent [8]	Chiang, H.-T., Wu, Y.-C., Yu, C., Toda, T., Wang, H.-M., Hu, Y.-C., \& Tsao, Y. (2021). HASA-Net: A Non-Intrusive Hearing-Aid Speech Assessment Network. \textit{2021 IEEE Automatic Speech Recognition and Understanding Workshop (ASRU)}, 907–913. \url{https://doi.org/10.1109/ASRU51503.2021.9687972}

\noindent [9]	Christensen, J. H., Pontoppidan, N. H., Rossing, R., Anisetti, M., Bamiou, D.-E., Spanoudakis, G., Murdin, L., Bibas, T., Kikidiks, D., Dimakopoulos, N., Giotis, G., \& Ecomomou, A. (2019a). Fully Synthetic Longitudinal Real-World Data From Hearing Aid Wearers for Public Health Policy Modeling. In \textit{Zenodo \url{https://doi.org/10.5281/zenodo.2668210}}.

\noindent [10]	Christensen, J. H., Pontoppidan, N. H., Rossing, R., Anisetti, M., Bamiou, D.-E., Spanoudakis, G., Murdin, L., Bibas, T., Kikidiks, D., Dimakopoulos, N., Giotis, G., \& Ecomomou, A. (2019b). Fully Synthetic Longitudinal Real-World Data From Hearing Aid Wearers for Public Health Policy Modeling. \textit{Front Neurosci, 13(850)}.

\noindent [11]	Garg, A. (2022). Speech enhancement using long short term memory with trained speech features and adaptive wiener filter. \textit{Multimedia Tools and Applications}. \url{https://doi.org/10.1007/s11042-022-13302-3}

\noindent [12]	Genolini, C., Écochard, R., \& Jacqmin-Gadda, H. (2013). Copy Mean: A New Method to Impute Intermittent Missing Values in Longitudinal Studies. \textit{Open Journal of Statistics, 03(04)}, 26–40. \url{https://doi.org/10.4236/ojs.2013.34A004}

\noindent [13]	Hochreiter, S., \& Schmidhuber, J. (1997). Long short-term memory. \textit{Neural Computation, 9(8)}, 1735–1780.

\noindent [14]	Houmøller, S. S., Wolff, A., Möller, S., Narne, V. K., Narayanan, S. K., Godballe, C., Hougaard, D. D., Loquet, G., Gaihede, M., Hammershøi, D., \& Schmidt, J. H. (2022). Prediction of successful hearing aid treatment in first-time and experienced hearing aid users: Using the International Outcome Inventory for Hearing Aids. \textit{International Journal of Audiology, 61(2)}, 119–129. \url{https://doi.org/10.1080/14992027.2021.1916632}

\noindent [15]	Iliadou, E., Su, Q., Kikidis, D., Bibas, T., \& Kloukinas, C. (2022). Profiling Hearing Aid Users through Big Data Explainable Artificial Intelligence Techniques. \textit{Frontiers in Neurology – Neuro-Otology}. \url{https://doi.org/10.3389/fneur.2022.933940}

\noindent [16]	Ilyas, I. F., \& Chu, X. (2019). Data cleaning. \textit{ACM}.

\noindent [17]	Iwahashi, J. H., de Souza Jardim, I., \& Bento, R. F. (2013). Results of hearing aids use dispensed by a publicly-funded health service. \textit{Brazilian Journal of Otorhinolaryngology, 79(6)}, 681–687. \url{https://doi.org/10.5935/1808-8694.20130126}

\noindent [18]	Jilla, A. M., Johnson, C. E., Danhauer, J. L., Anderson, M., Smith, J. N., Sullivan, J. C., \& Sanchez, K. R. (2020). Predictors of Hearing Aid Use in the Advanced Digital Era: An Investigation of Benefit, Satisfaction, and Self-Efficacy. \textit{Journal of the American Academy of Audiology, 31(2)}, 87–95. \url{https://doi.org/10.3766/jaaa.18036}

\noindent [19]	Kaplan-Neeman, R., Muchnik, C., Hildesheimer, M., \& Henkin, Y. (2012). Hearing aid satisfaction and use in the advanced digital era. \textit{The Laryngoscope, 122(9)}, 2029–2036. \url{https://doi.org/10.1002/lary.23404}

\noindent [20]	Kochkin, S. (2010). MarkeTrak VIII: Consumer satisfaction with hearing aids is slowly increasing. \textit{The Hearing Journal, 63(1)}, 19–20. \url{https://doi.org/10.1097/01.HJ.0000366912.40173.76}

\noindent [21]	Kuebert, T., Puder, H., \& Koeppl, H. (2021). Improving Daily Routine Recognition in Hearing Aids Using Sequence Learning. \textit{IEEE Access, 9}, 93237–93247. \url{https://doi.org/10.1109/ACCESS.2021.3092763}

\noindent [22]	Kuhn, M., Johnson, K., \& others. (2013). \textit{Applied predictive modeling (Vol. 26)}. Springer.

\noindent [23]	Lan, T., Ye, W., Lyu, Y., Zhang, J., \& Liu, Q. (2020). Embedding Encoder-Decoder With Attention Mechanism for Monaural Speech Enhancement. \textit{IEEE Access, 8}, 96677–96685. \url{https://doi.org/10.1109/ACCESS.2020.2995346}

\noindent [24]	Laplante-Lévesque, A., Kathleen Pichora-Fuller, M., \& Gagné, J.-P. (2006). Providing an internet-based audiological counselling programme to new hearing aid users: A qualitative study. \textit{International Journal of Audiology, 45(12)}, 697–706. \url{https://doi.org/10.1080/14992020600944408}

\noindent [25]	Laureyns, M., Bisgaard, N., Bobeldijk, M., \& Zimmer, S. (2020). Getting the numbers right on Hearing Loss Hearing Care and Hearing Aid Use in Europe, A Europe Wide Strategy Joint AEA, EFHOH, EHIMA report [Internet].

\noindent [26]	Lenatti, M., Moreno-Sánchez, P. A., Polo, E. M., Mollura, M., Barbieri, R., \& Paglialonga, A. (2022). Evaluation of Machine Learning Algorithms and Explainability Techniques to Detect Hearing Loss From a Speech-in-Noise Screening Test. \textit{American Journal of Audiology}, 1–19. \url{https://doi.org/10.1044/2022_AJA-21-00194}

\noindent [27]	Liu, G., \& Guo, J. (2019). Bidirectional LSTM with attention mechanism and convolutional layer for text classification. \textit{Neurocomputing}, 337, 325–338. \url{https://doi.org/10.1016/j.neucom.2019.01.078}

\noindent [28]	Lu, Y., Wang, M., Yao, L., Shen, H., Wu, W., Zhang, Q., Zhang, L., Chen, M., Liu, H., Peng, R., Liu, M., \& Chen, S. (2021). Auditory attention decoding from electroencephalography based on long short-term memory networks. \textit{Biomedical Signal Processing and Control, 70}, 102966. \url{https://doi.org/10.1016/j.bspc.2021.102966}

\noindent [29]	Lundberg, S. M., \& Lee, S.-I. (2017). A Unified Approach to Interpreting Model Predictions. Proceedings of the 31st \textit{International Conference on Neural Information Processing Systems}, 4768–4777.

\noindent [30]	Preeti, Bala, R., \& Singh, R. P. (2019). Financial and Non-Stationary Time Series Forecasting using LSTM Recurrent Neural Network for Short and Long Horizon. \textit{2019 10th International Conference on Computing, Communication and Networking Technologies (ICCCNT)}, 1–7. \url{https://doi.org/10.1109/ICCCNT45670.2019.8944624}

\noindent [31]	Robertson, M. A., Kelly-Campbel, R. J., \& Wark, D. J. (2012). Use of Audiometric Variables to Differentiate Groups of Adults Based on Hearing Aid Ownership and Use. \textit{Contemporary Issues in Communication Science and Disorders, 39(Fall)}, 114–120. \url{https://doi.org/10.1044/cicsd_39_F_114}

\noindent [32]	Rosenfeld, A. (2021). Better Metrics for Evaluating Explainable Artificial Intelligence. \textit{20th International Foundation for Autonomous Agents and Multiagent Systems (AAMAS ’21)}, 45–50.

\noindent [33]	Singh, G., Lau, S.-T., \& Pichora-Fuller, M. K. (2015). Social Support Predicts Hearing Aid Satisfaction. \textit{Ear \& Hearing}, 36(6), 664–676. \url{https://doi.org/10.1097/AUD.0000000000000182}

\noindent [34]	Tjoa, E., \& Guan, C. (2021). A Survey on Explainable Artificial Intelligence (XAI): Toward Medical XAI. \textit{IEEE Transactions on Neural Networks and Learning Systems, 32(11)}, 4793–4813. \url{https://doi.org/10.1109/TNNLS.2020.3027314}

\noindent [35]	Tomiazzi, J. S., Pereira, D. R., Judai, M. A., Antunes, P. A., \& Favareto, A. P. A. (2019). Performance of machine-learning algorithms to pattern recognition and classification of hearing impairment in Brazilian farmers exposed to pesticide and/or cigarette smoke. \textit{Environmental Science and Pollution Research, 26(7)}, 6481–6491. \url{https://doi.org/10.1007/s11356-018-04106-w}

\noindent [36]	Vaswani, A., Shazeer, N., Parmar, N., Uszkoreit, J., Jones, L., Gomez, A. N., Kaiser, L., \& Polosukhin, I. (2017). Attention Is All You Need.

\noindent [37]	World Health Organization. (2017). Global costs of unaddressed hearing loss and cost-effectiveness of interventions: a WHO report, 2017. World Health Organization.

\noindent [38]	Xu, K., Ba, J., Kiros, R., Cho, K., Courville, A., Salakhudinov, R., Zemel, R., \& Bengio, Y. (2015). Show, Attend and Tell: Neural Image Caption Generation with Visual Attention. In F. Bach \& D. Blei (Eds.), \textit{Proceedings of Machine Learning Research} (pp. 2048–2057). PMLR.

\noindent [39]	Zhang, Z., Shen, Y., \& Williamson, D. S. (2019). Objective Comparison of Speech Enhancement Algorithms with Hearing Loss Simulation. \textit{ICASSP 2019 - 2019 IEEE International Conference on Acoustics, Speech and Signal Processing (ICASSP)}, 6845–6849. \url{https://doi.org/10.1109/ICASSP.2019.8683040}

\noindent [40]	Zhao, Y., Li, J., Zhang, M., Lu, Y., Xie, H., Tian, Y., \& Qiu, W. (2019). Machine Learning Models for the Hearing Impairment Prediction in Workers Exposed to Complex Industrial Noise: A Pilot Study. \textit{Ear \& Hearing, 40(3)}, 690–699. \url{https://doi.org/10.1097/AUD.0000000000000649}
 
\clearpage

\end{document}